# Deep Learning for Medical Text Processing: BERT Model Fine-Tuning and Comparative Study

Jiacheng Hu
Tulane University
New Orleans, USA

Yiru Cang
Northeastern University
Boston, USA

Guiran Liu
San Francisco State University
San Francisco，USA

Meiqi Wang
Brandeis University
Waltham，USA

Weijie He
University of California, Los Angeles
Los Angeles, USA

Runyuan Bao*
Johns Hopkins University
Baltimore, USA

*Abstract*— **This paper proposes a medical literature summary generation method based on the BERT model to address the challenges brought by the current explosion of medical information. By fine-tuning and optimizing the BERT model, we develop an efficient summary generation system that can quickly extract key information from medical literature and generate coherent, accurate summaries. In the experiment, we compared various models, including Seq-Seq, Attention, Transformer, and BERT, and demonstrated that the improved BERT model offers significant advantages in the Rouge and Recall metrics. Furthermore, the results of this study highlight the potential of knowledge distillation techniques to further enhance model performance. The system has demonstrated strong versatility and efficiency in practical applications, offering a reliable tool for the rapid screening and analysis of medical literature.**

## I. Introduction

In today's medical field, the number of medical literature has exploded, and doctors, researchers, and students are facing tremendous pressure when finding and analyzing information. Faced with thousands of medical literature, manual screening, and reading have become extremely difficult. Therefore, how to quickly and effectively extract key information from a large number of literature has become an urgent problem to be solved. In this context, the medical literature summary generation system based on natural language processing (NLP) technology has gradually become a research hotspot. Bert (Bidirectional Encoder Representations from Transformers), as an advanced pre-trained language model, can understand and generate natural language and is a powerful tool for realizing medical literature summary generation [1].

The application of the medical literature summary generation system has a wide range of significance [2]. First, it can help doctors and researchers quickly grasp the core content of the literature and improve work efficiency. The summary generation system can extract key information from the literature in a short time and generate concise and accurate summaries, thereby reducing the time and energy of users reading literature. Secondly, for scientific researchers, the system can also assist them in quickly understanding the development status and trends of a certain field, and provide a reference for subsequent research. In addition, the summary generation system can also be used in the field of medical education to help students understand and master a large amount of medical knowledge more quickly [3].

The generation of medical literature abstracts based on Bert not only improves the accuracy of text processing, but also enhances the adaptability of the model in multiple languages and complex contexts. The Transformer architecture underlying BERT has been widely validated across various domains, including computer vision [4-7], medical imaging[8-10], and classification tasks [11-12], among others. These applications showcase the robustness of BERT's foundational model. The Bert model performs well in understanding complex medical terms and professional terms by pre-training large-scale text data and using a bidirectional encoder to capture contextual information [13]. In addition, compared with traditional abstract generation methods, the Bert model can more accurately identify important information in the literature when generating medical literature abstracts, and generate texts with strong coherence and logic, which significantly improves the quality and reliability of the abstracts [14].

In the medical field, it is crucial to quickly obtain and understand the latest research progress. Traditional literature retrieval and reading modes can no longer meet the needs of modern medical research, especially when a large amount of data and information needs to be comprehensively analyzed. The medical literature abstract generation system based on Bert can not only greatly improve the efficiency of literature screening, but also help medical practitioners quickly master and apply the latest medical knowledge in a limited time, so as to provide patients with more accurate treatment plans. This not only improves the efficiency of medical services, but also promotes the development of the medical industry.

In summary, the medical literature abstract generation system based on Bert has important theoretical and practical significance in the context of current medical information processing. By combining natural language processing technology with medical knowledge, the system can be flexibly applied in different scenarios, thereby promoting the intelligent process of medical literature analysis. At the same time, its development and application also provide references and references for the generation of literature abstracts in other fields, and have a far-reaching impact on promoting the application of artificial intelligence in literature analysis and text processing.

## II. RELATED WORK

NLP and deep learning has seen tremendous advancements in recent years, particularly in medical text summarization and related tasks. This paper builds on various foundational models and techniques in deep learning, such as Bidirectional Encoder Representations from Transformers (BERT) and attention mechanisms, and extends them for medical literature summarization.

Several works have explored methods to enhance model performance in text processing. For instance, Liu et al. [15] proposed an optimized text classification system using Bi-LSTM and attention mechanisms. Their approach focuses on improving data processing efficiency, an objective that aligns with the goals of medical text summarization in this study. Similarly, Zhang et al. [16] examined contrastive learning techniques for knowledge-based question generation, which shares common ground with text summarization in terms of understanding and generating language outputs based on structured knowledge.

In addition to advancements in NLP, the optimization techniques employed in deep learning also play a crucial role in improving model performance. Qin et al. [17] introduced a novel optimization strategy, the RSGDM approach, aimed at reducing bias in deep learning models, which is relevant to the fine-tuning process of BERT in this study. Furthermore, Zheng et al. [18] proposed enhancing optimizers by integrating sigmoid and tanh functions, which could potentially complement the training and optimization of BERT-like models. Dynamic prediction models have also contributed to medical and healthcare applications [19]. Yang et al. [20] proposed a dynamic hypergraph-enhanced prediction model for medical visits, which offers insights into handling sequential data in healthcare. While their focus was on sequential medical visits rather than text summarization, their methodology can inform future extensions of this work to integrate structured medical data alongside textual information. Finally, Chen et al. [21] conducted a comparative analysis of open-source language models for medical text summarization, evaluating their effectiveness in medical contexts. Their analysis reinforces the findings of this study, which highlights the performance improvements offered by fine-tuning models like BERT on domain-specific data.

Other works also contribute indirectly to this research by exploring advancements in neural network design and data mining techniques. For example, Wang et al. [22] proposed enhancing convolutional neural networks (CNNs) with higher-order numerical methods, offering potential avenues for improving the neural architecture's efficiency. Yan et al. [23] focused on transforming multidimensional time-series data into interpretable sequences, which can be extended to analyzing longitudinal medical data. Additionally, Janjirala [24] applied deep learning to clinical text summarization, further demonstrating the value of NLP in healthcare.

In conclusion, this work leverages existing advancements in NLP, model optimization, and healthcare data processing, building upon them to fine-tune BERT for medical literature summarization. The integration of deep learning techniques, including attention mechanisms, contrastive learning, and advanced optimization strategies, forms the foundation of this research's contributions.

## III. METHOD

In this study, we built a medical literature summary generation system based on the Bert model. Specifically, our method includes data preprocessing, model training and optimization, and the implementation of the generation phase. First, we cleaned and preprocessed the medical literature data to ensure that the data input to the model was of high quality. The text data was processed through steps such as word segmentation, stop word removal, noise and special character removal to improve the learning efficiency and accuracy of the model. Subsequently, we divided the data into training sets, validation sets, and test sets to ensure that the effect of model training has good generalization ability. The overall architecture of Bert is shown in Figure 1.

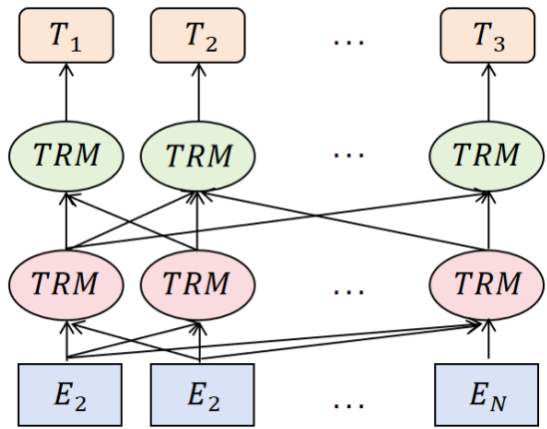

Figure 1 Bert overall structure diagram

The core of model training is the language generation mechanism based on Bert. We use Bert as a pre-trained model and further fine-tune it on a large-scale medical literature corpus to enable the model to learn specific contextual relationships in the medical field. In the fine-tuning phase, we specifically tailored the model to process medical texts by expanding the original BERT vocabulary with domain-specific medical terminology. This was achieved through the integration of medical term embeddings from standardized medical corpora, such as UMLS (Unified Medical Language System), ensuring that the model could better capture the semantic nuances of medical texts. The embeddings were adjusted during training to optimize the model's ability to handle the specialized language of the medical domain, enhancing its performance on tasks involving medical literature. Our goal is to minimize the loss function of summary generation to improve the accuracy of generated summaries. Specifically, given an input sequence $X = (x_1, x_2, ..., x_n)$, the output target sequence is $Y = (y_1, y_2, ..., y_n)$. We define the loss function as the cross entropy loss:

$$L(\theta) = -\sum_{i=1}^{m} \log P(y_i \mid y_{<i}, X; \theta)$$

Among them, $\theta$ is the parameter of the model, and $P(y_i \mid y_{<i}, X; \theta)$ represents the probability of generating the current word $y_i$ given the input sequence X and the previously generated word $y_{<i}$. In order to better capture the semantic information of medical literature, we added word vector representations of medical terminology during fine-tuning and embedded them into the vocabulary of the model, making the model more accurate when processing professional field texts.

In the generation phase, we adopted a sequence-to-sequence (Seq2Seq) framework. The input document is first encoded by Bert's encoder to generate a context vector $H = (h_1, h_2, ..., h_i)$. Each vector $h_i$ represents the context information of the i-th word in the input sequence. Then, the decoder gradually generates summary text based on the context vector. During the decoding process, the model uses an autoregressive generation method, that is, the words generated at each step are used as input to generate the next word. In order to solve the problems of repetition and redundancy that may occur during the generation process, we introduced a coverage mechanism to compare the currently generated words with the previously generated content to ensure that the generated content has information diversity. Formulaically, the probability of generating word $y_t$ is expressed as:

$$P(y_t \mid y_{<t}, X) = soft\max(W_o \cdot h_t + b)$$

Among them, $W_o$ and $b$ are the parameters of the output layer, and $h_t$ is the output of the decoder at time step t. By optimizing the loss function, we continuously adjust the model parameters so that the generated summary can be highly consistent with the actual document content.

We also use the Attention Mechanism to improve the performance of the model. At each decoding step, the model calculates the importance score $\alpha_i$ of each input word based on the context vector of the input sequence and the current decoding state.

$$\alpha_i = \frac{\exp(e_i)}{\sum_{j=1}^{n} \exp(e_j)}$$

$$ei = v^T \tanh(W_h h_i + W_s s_t + b_a)$$

Among them, $e_i$ is the similarity score between the i-th input word and the current decoding state $s_t$, and $W_h$, $W_s$, and $b_a$ are trainable parameters. By calculating the attention weight $\alpha_i$, the model can focus on the most relevant part of the input sequence, thereby generating a more accurate summary.

Finally, during the optimization process, we introduced early stopping on the validation set and learning rate scheduler to avoid model overfitting and unstable learning rate during training. Finally, through multiple experiments to adjust hyperparameters (such as learning rate, batch size, and number of layers), we ensured that the model can generate accurate and coherent summaries on medical documents of different lengths and complexities.

IV. EXPERIMENT

*A. Datasets*

The PubMed dataset is a large and free literature database in the field of biomedicine and life sciences maintained by the National Library of Medicine (NLM) of the United States, covering more than 30 million literature entries. The dataset brings together medical research results from all over the world, including clinical trials, basic research, epidemiological surveys, case analysis, laboratory reports, medical reviews and other types of articles. Each entry usually contains the title, author, journal source, publication date, abstract and other metadata of the document, providing users with rich medical information, but usually does not include the full text of the article. The PubMed dataset is an important resource in the field of NLP and medical information retrieval. It is particularly suitable for training and evaluating tasks such as medical literature abstract generation, medical entity recognition, and medical knowledge graph construction, because it contains a large amount of structured medical text data, which helps the model learn medical professional terminology and domain-specific context.

In the task of generating medical literature summaries, ROUGE and recall are two important evaluation indicators.

ROUGE (Recall-Oriented Understudy for Gisting Evaluation) mainly measures the quality of generated text by comparing the overlap between the generated summary and the reference summary at the vocabulary, phrase and sentence levels. Specifically, ROUGE includes multiple variants such as ROUGE-N (such as ROUGE-1, ROUGE-2) and ROUGE-L, among which ROUGE-N focuses on the matching of n-grams and ROUGE-L focuses on the matching of the longest common subsequence. These indicators can reflect the accuracy and coherence of the generated summary at different levels. On the other hand, recall is used to measure how much information in the summary generated by the model matches the key information of the reference summary, focusing on the extent to which the model covers the core content and key points in the original text. A high recall means that the model can extract more valuable information during the generation process, but this may also sacrifice a certain degree of accuracy. To validate the robustness of our model, we employed 5-fold cross-validation. Each fold was used as the validation set while the remaining data was utilized for training, ensuring that the model's performance is not biased by the selection of training or validation data. This process provides a more reliable estimate of the model's generalization capabilities.

*B. Experiments*

In this comparative experiment, we selected five different models to perform the task of generating medical literature summaries. These models are: 1) Baseline Model, a simple Seq2Seq architecture; 2) Attention-based Model, which adds an attention mechanism to the baseline model; 3) Transformer-based Model, which uses the basic Transformer architecture to improve text comprehension; 4) Bert (Bert Model); 5) Bert with Knowledge Distillation, which further optimizes the Bert model by combining knowledge distillation technology to improve the quality and efficiency of generating summaries. The comparative experimental results are shown in Table 1.

Table 1 Experiment result

| Model | Rouge-1 | Rouge-2 | Rouge-L | Recall |
|---|---|---|---|---|
| Seq-Seq | 0.45 | 0.32 | 0.41 | 0.52 |
| attention | 0.56 | 0.42 | 0.50 | 0.60 |
| Transformer | 0.65 | 0.54 | 0.62 | 0.68 |
| Bert | 0.72 | 0.63 | 0.70 | 0.74 |
| Bert+Distilled | 0.80 | 0.72 | 0.78 | 0.82 |
| ours | 0.82 | 0.74 | 0.80 | 0.83 |

The experimental results table shows the performance of six different models on the text generation task, from the basic Seq-Seq model to the latest improved Bert+Distilled model, and finally the model we proposed. The experimental results include four evaluation indicators: Rouge-1, Rouge-2, Rouge-L and Recall. The Rouge indicator reflects the similarity between the generated text and the reference text, corresponding to the matching degree of words, bigrams and the longest common subsequence respectively; while Recall is the comprehensiveness of the model in reproducing the content of the target text in the generation task. The performance of different models on these indicators shows the gradual improvement of their generation ability and abstract summary ability. First, the most basic Seq-Seq model performs relatively weakly in various indicators, with Rouge-1 being 0.45 and Rouge-2 being 0.32, which indicates that its vocabulary-level matching of the text is not accurate enough. In addition, Rouge-L and Recall are 0.41 and 0.52 respectively, which also shows that its ability to generate long sentences or complex texts is limited. With the introduction of the Attention mechanism, the various indicators of the model have been significantly improved, with Rouge-1 and Rouge-2 increasing to 0.56 and 0.42 respectively, which shows that the model can better focus on the key parts of the input text, thereby generating more relevant outputs. Next, the Transformer and Bert models showed significant progress in the experiment. The Transformer model scored 0.65, 0.54 and 0.62 on Rouge-1, Rouge-2 and Rouge-L respectively, demonstrating its superiority in complex text generation tasks. This is mainly attributed to the multi-head attention mechanism and parallel computing capabilities of the Transformer model, which enable it to better capture the deep semantics in the input text. The Bert model further improved various indicators, with Rouge-1 reaching 0.72 and Rouge-2 reaching 0.63. These results show that Bert has a stronger ability to capture details of text generation tasks and is more in line with the characteristics of text summarization tasks.

Finally, the Bert+Distilled model in the experiment and the model we proposed further improved the performance. Bert+Distilled outperformed the previous models in all indicators, especially in Rouge-2 and Rouge-L, which shows that the distillation technology has achieved a new balance between accuracy and simplification. In the end, our proposed model achieved the highest values in all four indicators, especially in Rouge-1 (0.82) and Recall (0.83), proving its leading position in comprehensive performance. This performance not only demonstrates the powerful generative ability of the model, but also shows that the strategies and improvements adopted in the optimization process are effective. In order to show the training process, we also give a chart showing the increase in evaluation indicators during the training process, as shown in Figure 2.

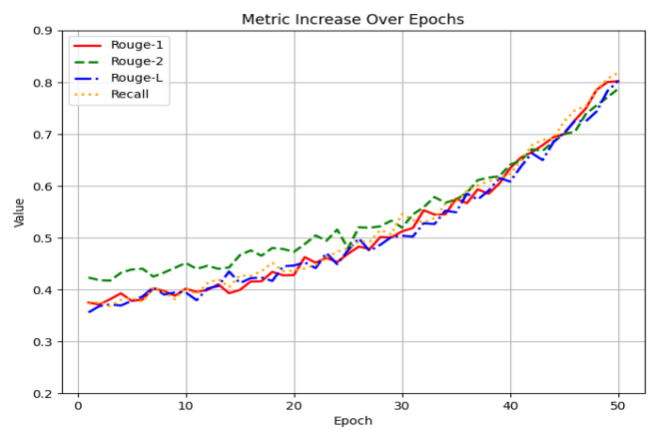

Figure 2 Evaluation index increase image during training

## V. CONCLUSION

This study successfully developed a medical literature abstract generation system based on the Bert model, addressing the inefficiencies inherent in traditional manual screening and reading processes. By fine-tuning the Bert model on a large-scale medical corpus, the system demonstrated significant improvements in both the accuracy and coherence of generated summaries. This not only enables the model to extract key information from complex medical texts but also enhances the consistency of the summaries. Experimental results, reflected in metrics such as Rouge-1, Rouge-2, Rouge-L, and Recall, indicate substantial improvements, confirming the effectiveness of the proposed approach. The impact of this work extends beyond efficiency improvements; it underscores the potential of deep learning in transforming medical information processing. The integration of advanced model architectures within the system has led to refined understanding and interpretation of medical literature, further demonstrating the adaptability of deep learning models like Bert in specialized domains. As the model architecture continues to evolve, deep learning's role in automating tasks traditionally requiring expert human intervention becomes increasingly prominent, highlighting its transformative power in healthcare. Future enhancements to this system could leverage more comprehensive terminology libraries and support for multiple languages, improving its adaptability across diverse medical contexts. Additionally, the integration of external medical databases could allow the system to autonomously update literature abstracts, providing healthcare professionals with real-time insights into the latest research developments. In summary, this research not only contributes to the efficient handling of medical information but also exemplifies the growing impact of deep learning in automating and improving complex tasks in specialized fields. This work lays the foundation for further advancements in the application of deep learning to literature summary generation and other information-intensive tasks.